\documentclass[10pt,twocolumn,letterpaper]{article}

\usepackage[final]{cvpr}      

\usepackage{graphicx}
\usepackage{amsmath}
\usepackage{amssymb}
\usepackage{booktabs}

\usepackage{multirow}
\usepackage{sidecap}
\usepackage{float}
\usepackage{bm}
\usepackage{algorithmic}
\usepackage[ruled]{algorithm2e}
\SetKwInput{KwInput}{Input}
\SetKwInput{KwOutput}{Output}

\usepackage[pagebackref,breaklinks,colorlinks]{hyperref}

\usepackage[capitalize]{cleveref}
\crefname{section}{Sec.}{Secs.}
\Crefname{section}{Section}{Sections}
\Crefname{table}{Table}{Tables}
\crefname{table}{Tab.}{Tabs.}

\def\cvprPaperID{*****} 
\def\confName{****}
\def\confYear{****}

\begin{document}
	
	\title{Robust and Accurate Object Detection via Self-Knowledge Distillation}
	
	\author{
		Weipeng Xu\textsuperscript{1}
		\enskip Pengzhi Chu\textsuperscript{1}
		\enskip Renhao Xie\textsuperscript{1}
		\enskip Xiongziyan Xiao\textsuperscript{1}
		\enskip Hongcheng Huang\textsuperscript{1}\thanks{Corresponding author.}\\ \vspace{-0.3em}
		\\
		\textsuperscript{1}Shanghai Jiao Tong University \\
	}
	\maketitle
	
	\begin{abstract}
		Object detection has achieved promising performance on clean datasets, but how to achieve better tradeoff between the adversarial robustness and clean precision is still under-explored. Adversarial training is the mainstream method to improve robustness, but most of the works will sacrifice clean precision to gain robustness than standard training. In this paper, we propose Unified Decoupled Feature Alignment (UDFA), a novel fine-tuning paradigm which achieves better performance than existing methods, by fully exploring the combination between self-knowledge distillation and adversarial training for object detection. We first use decoupled fore/back-ground features to construct self-knowledge distillation branch between clean feature representation from pretrained detector (served as teacher) and adversarial feature representation from student detector. Then we explore the self-knowledge distillation from a new angle by decoupling original branch into a self-supervised learning branch and a new self-knowledge distillation branch. With extensive experiments on the PASCAL-VOC and MS-COCO benchmarks, the evaluation results show that UDFA can surpass the standard training and state-of-the-art adversarial training methods for object detection. For example, compared with teacher detector, our approach on GFLV2 with ResNet-50 improves clean precision by 2.2 AP on PASCAL-VOC; compared with SOTA adversarial training methods, our approach improves clean precision by 1.6 AP, while improving adversarial robustness by 0.5 AP. Our code will be available at https://github.com/grispeut/udfa.

	\end{abstract}

	\section{Introduction}
	\label{sec:intro}
	\begin{figure}[!htb]
		\centering
		\includegraphics[width=0.95\columnwidth]{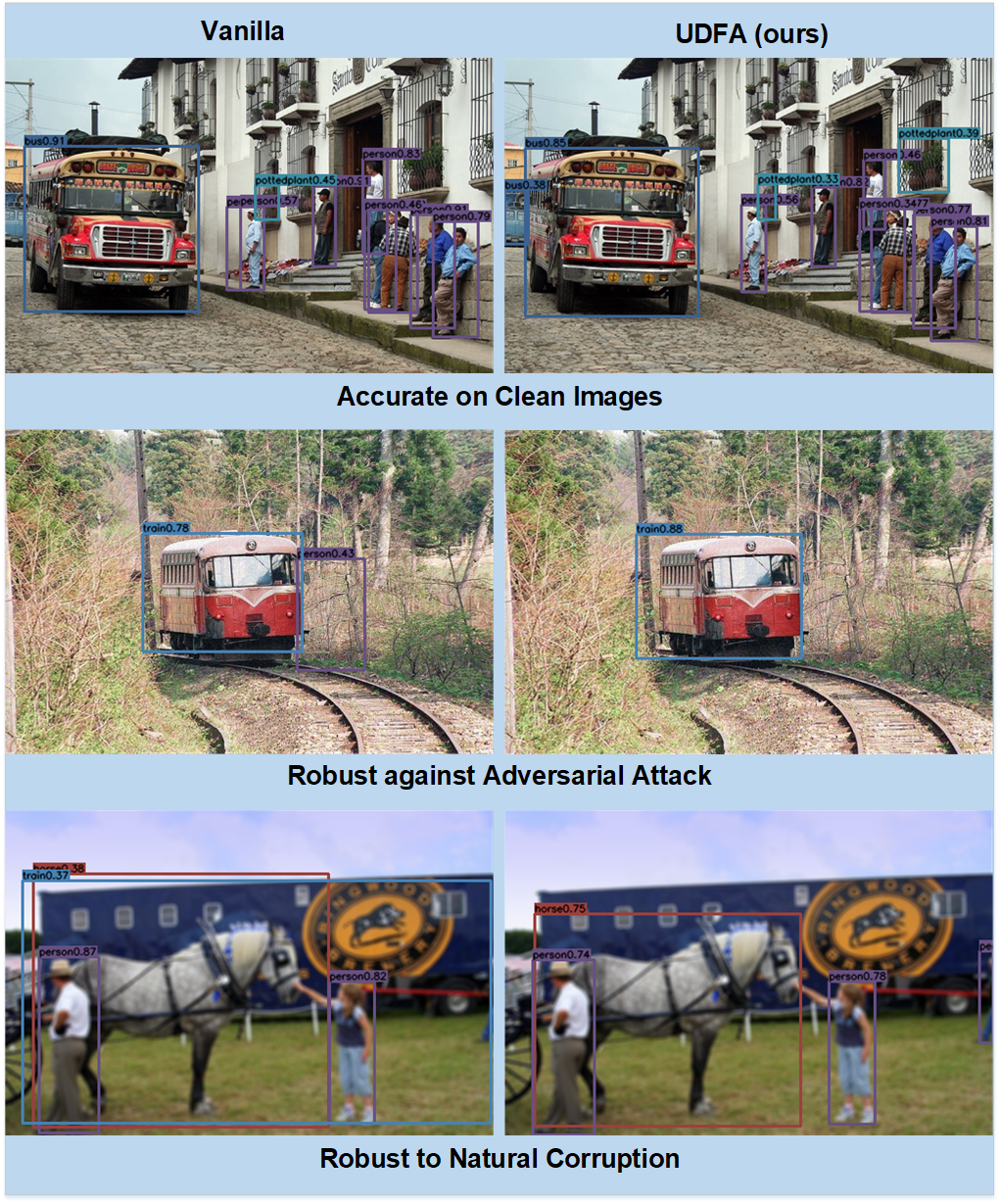}
		\caption{ \textbf{Vanilla (standard training) v.s. UDFA} on different settings.
			\textbf{Top:}
			For clean images, UDFA can correctly detect some objects (e.g., ``pottedplant" and ``bus") missed by the vanilla detector, especially in an image which exists multiple objects.
			\textbf{Middle:}
			For adversarial images, UDFA can reduce false detection result (e.g., ``person") and increase the confidence of true detection result (e.g., ``train"). 
			\textbf{Bottom:}
			For natural corrupted images, UDFA generates more precise location for ``horse" and the vanilla detector produces a false positive for ``train” after the image is corrupted by blur. \textbf{Zoom in for more visual details.}}
		\vspace{-12pt}
		\label{fig:images}
	\end{figure}

	Deep learning has made a significant breakthrough in many computer vision tasks such as classification \cite{DBLP:conf/cvpr/HeZRS16,DBLP:journals/pami/GaoCZZYT21,DBLP:conf/iclr/DosovitskiyB0WZ21}, object detection \cite{DBLP:conf/cvpr/LinDGHHB17,DBLP:conf/nips/0041WW00LT020,DBLP:conf/cvpr/LiW0LT021} and instance segmentation \cite{DBLP:conf/iccv/HeGDG17,DBLP:conf/eccv/ChenZPSA18}. However, adversarial samples \cite{DBLP:journals/corr/SzegedyZSBEGF13,DBLP:journals/corr/GoodfellowSS14,DBLP:conf/iccv/XieWZZXY17,DBLP:conf/icml/Croce020a} are still viewed as a troublesome threat to deep learning models. Adversarial samples are crafted for leading deep learning models to make wrong predictions by adding imperceptible perturbations to clean images. Generally, when the knowledge of structures and parameters of a given model is available, various methods can successfully fool the model with so called white-box attack \cite{DBLP:journals/corr/GoodfellowSS14,DBLP:conf/iclr/MadryMSTV18,DBLP:conf/iccv/XieWZZXY17,DBLP:conf/bmvc/LiTCBL18}. 
	
	In order to improve the model robustness, many efforts have been made to defend adversarial attack \cite{DBLP:journals/corr/GoodfellowSS14,DBLP:conf/sp/PapernotM0JS16,DBLP:conf/iclr/MetzenGFB17,DBLP:conf/iclr/MadryMSTV18,DBLP:conf/iclr/Ma0WEWSSHB18,DBLP:conf/icml/WangM0YZG19}. Among them, adversarial training \cite{DBLP:journals/corr/GoodfellowSS14,DBLP:conf/iclr/MadryMSTV18} has been demonstrated to be the most effective method so far. Based on adversarial training, a number of approaches have been proposed to enhance its performance further. For example, AWP \cite{DBLP:conf/nips/WuX020} forms a double-perturbation mechanism in the adversarial training framework; TRADES \cite{DBLP:conf/icml/ZhangYJXGJ19} optimizes an upper bound of adversarial risk to balance accuracy and robustness; MART \cite{DBLP:conf/iclr/0001ZY0MG20} adds an extra regularization of misclassified examples in adversarial training. Although the above approaches have verified their effectiveness on classification task, they do not extend their work to the object detection task. Compared with classification 
	task, detection requires predicting the locations of semantic objects, which is complicated but also important.
	
	
	To improve the robustness of object detection models, Zhang et al. \cite{DBLP:conf/iccv/ZhangW19} first present a method to combine adversarial training with detection tasks by using task oriented domain to generate adversarial samples in adversarial training (denoted as TOD-AT).  Chen et al. \cite{DBLP:conf/cvpr/ChenKC21} further propose class-wise adversarial training especially for the situations where multiple objects of different classes exist in single image. The above two works can indeed enhance adversarial robustness of detectors than standard training. However, performance on clean datasets drops significantly (about a drop of 20\% mAP on PASCAL-VOC \cite{DBLP:journals/ijcv/EveringhamGWWZ10}), which is not conductive to real-world applications. In order to balance clean precision and robustness, Feature Alignment \cite{9506689} proposes to guide the outputs of middle layers for better adversarial training (denoted as Vanilla-FA), which builds on adversarial training, knowledge distillation, and self-supervised learning. Although Vanilla-FA can mitigate the drop of clean precision and boost robustness of detectors, the clean precision is still lower than standard training. For robust and accurate object detection, Det-AdvProp \cite{DBLP:conf/cvpr/ChenXTZHG21} combines TOD-AT with AdvProp \cite{DBLP:conf/cvpr/XieTGWYL20}, making clean precision and robustness outperform standard training. To the best of our knowledge, Det-AdvProp is the state-of-the-art (SOTA) result in adversarial-training-based algorithms for object detection, which does not sacrifice clean accuracy to gain robustness. However, on PASCAL-VOC benchmark, we find that Det-AdvProp fails to obtain improvements over pretrained detector on clean precision when we apply corrupted augmentation \cite{DBLP:journals/corr/abs-1907-07484} (an effective way to boost natural corrupted robustness) during training.

	This paper aims at building more accurate and robust detectors. Motivated by Vanilla-FA, we propose a Unified Decoupled Feature Alignment (UDFA) framework, which can largely improve the performance than Vanilla-FA. Besides, UDFA outperforms standard training (shown in Figure \ref{fig:images}) and Det-AdvProp in terms of clean precision and robustness. Vanilla-FA uses additional Knowledge-Distilled Feature Alignment (KDFA) branch and Self-Supervised Feature Alignment (SSFA) branch to promote adversarial training. KDFA constructs contrastive loss between student and teacher; SSFA constructs contrastive loss between adversarial images and clean images from student. However, the interactions and dependencies of two branches have not been explored. They directly push student to imitate different feature representations from knowledge distillation and self-supervised learning, which will result in conflicts or competitions. We conjecture that the inner compromise leads to the sub-optimal results. Potential conflict in Vanilla-FA will be analyzed in Section \ref{sec:conflict}. In light of the lessons above, we propose to unify the two independent branches by knowledge distillation. We find that original KDFA branch can essentially be viewed as a self-supervised learning branch and a new knowledge distillation branch (described in Section \ref{sec:theory}). Therefore, our UDFA framework is composed of an original KDFA branch and two decoupled sub-branches, which can address the conflict in theory. This paper aims at exploring a novel fine-tuning paradigm by self-knowledge distillation, so we use pretrained detector as teacher.  In addition, we adopt DeFeat \cite{DBLP:conf/cvpr/Guo00W0X021} to fit it into the feature alignment framework. We show that decoupled features can improve the performance than global pooling feature used in Vanilla-FA. With extensive experiments on PASCAL-VOC \cite{DBLP:journals/ijcv/EveringhamGWWZ10} and MS-COCO \cite{DBLP:conf/eccv/LinMBHPRDZ14} benchmarks, the results demonstrate that our UDFA can effectively improve the clean accuracy and robustness than SOTA algorithms. To push the limit of our approach, we also verify the effectiveness of UDFA on natural corrupted datasets. The main contributions of our work can be summarized as follows:
	\begin{itemize}
		\item To our best knowledge, our approach is the first distillation-based approach which does not sacrifice clean precision to gain robustness in adversarial training. This work demonstrates the strength of knowledge distillation in the field of adversarial training. 
		\item We explain the self-knowledge distillation in adversarial training from a new perspective by viewing self-supervised learning as a sub-branch of original self-knowledge distillation branch. Then we propose UDFA, a unified framework, which surpasses the SOTA adversarial-training-based methods on PASCAL-VOC and MS-COCO.
		

		
	\end{itemize}

	\begin{figure*}[!htb]
		\centering
		
		\includegraphics[width=0.8\linewidth]{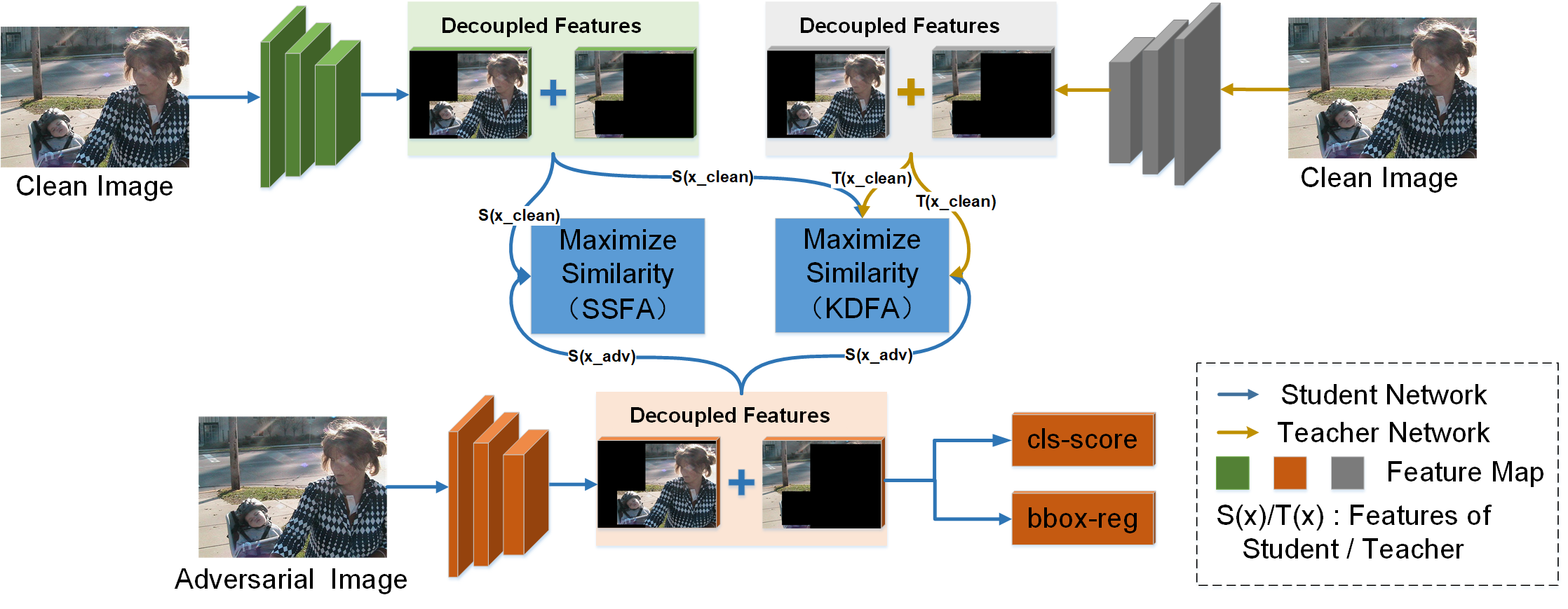}
		
		\caption{The overall pipeline of UDFA. The SSFA module constructs feature loss from student detector, and the KDFA module constructs feature loss between teacher and student. Several feature alignment branches can be viewed as a unified self-distillation process.  
		}
		\label{fig:overall_framework}
	\end{figure*}
	
	\section{Related Work}
	\paragraph{Adversarial Attack.}
	Adversarial attack for classification and object detection has been extensively studied recently \cite{DBLP:conf/iccv/XieWZZXY17,DBLP:journals/corr/GoodfellowSS14,DBLP:conf/iclr/MadryMSTV18,DBLP:conf/bmvc/LiTCBL18,DBLP:conf/iclr/Huang020,DBLP:conf/cvpr/LiXZYL20,DBLP:conf/cvpr/MahoFM21,DBLP:journals/tip/LiC21}. Existing approaches in adversarial attack can be categorized into two groups: white-box attack \cite{DBLP:journals/corr/GoodfellowSS14,DBLP:conf/iclr/MadryMSTV18,DBLP:conf/iccv/XieWZZXY17,DBLP:conf/bmvc/LiTCBL18} and black-box attack \cite{DBLP:conf/cvpr/LiXZYL20,DBLP:journals/tip/LiC21,DBLP:conf/iclr/Huang020,DBLP:conf/cvpr/MahoFM21}. White-box attack is more effective in testing the robustness of deep learning models, and black-box attack is more important in testing the transferability and strength of adversarial attack. Projected gradient descent (PGD) \cite{DBLP:conf/iclr/MadryMSTV18} is a powerful white-box attack when network structures and parameters are available. 
	
	\vspace{-10pt}
	\paragraph{Adversarial Training in Object Detection.}
	Adversarial training \cite{DBLP:journals/corr/GoodfellowSS14,DBLP:conf/iclr/MadryMSTV18} is mainstream method to improve adversarial robustness of deep learning models. For the robustness of object detector, Zhang et al. \cite{DBLP:conf/iccv/ZhangW19} take an initial step towards adversarially robust detector training. They decompose adversarial training into classification and localization domains to fit it into detection tasks. \cite{DBLP:conf/cvpr/ChenKC21} proposes class-wise adversarial training by decomposing the overall loss into class-wise losses. \cite{9506689} proposes Feature Alignment  to balance clean precision and robustness, which uses extra knowledge distillation branch and self-supervised learning branch in adversarial training. The above works all sacrifice clean accuracy more or less in adversarial training than standard training. Det-AdvProp \cite{DBLP:conf/cvpr/ChenXTZHG21} extends AdvProp 
	\cite{DBLP:conf/cvpr/XieTGWYL20} to the the scenario of object detection, which can get higher clean precision and robustness than standard training and AutoAugment \cite{DBLP:conf/cvpr/CubukZMVL19}.  
	
	\vspace{-10pt}
	\paragraph{Knowledge Distillation and Self-Supervised Learning in Object Detection.} Knowledge distillation has been widely adopted in model compression \cite{DBLP:journals/corr/HintonVD15,DBLP:journals/corr/RomeroBKCGB14,DBLP:conf/iccv/LiuLSHYZ17}. For compressing detection models, \cite{DBLP:conf/nips/ChenCYHC17} proposes to distill all components across neck features and detection heads. \cite{DBLP:conf/cvpr/LiJY17} proposes to distill the feature knowledge from region proposals. \cite{DBLP:conf/cvpr/WangYZF19} proposes to distill fine-grained features from foreground object regions. \cite{DBLP:conf/cvpr/Guo00W0X021} proposes to decouple the fore/back-ground features during knowledge distillation. Self-supervised contrastive learning has achieved
	SOTA performance in image classification \cite{DBLP:conf/cvpr/He0WXG20,DBLP:conf/icml/ChenK0H20,DBLP:conf/cvpr/ChenH21}. Recently, some works focus on advancing self-supervised learning for object detection. Instance Localization \cite{DBLP:conf/cvpr/YangWZL21} proposes to learn region-level representations by copy-pasting instances onto background images. DetCo \cite{DBLP:journals/corr/abs-2102-04803} proposes to build contrastive loss across the global and local representations.

	\section{Methods}
	\label{sec:app}
	
	In this section, we first analyze the potential drawbacks in Vanilla-FA \cite{9506689}, and then propose decoupled knowledge-distilled feature alignment branches to tackle the conflict. Finally, we introduce UDFA shown
	in Figure \ref{fig:overall_framework}. 
	
	\subsection{Potential Conflict in Vanilla-FA} 
	\label{sec:conflict}
	
	Vanilla-FA proposes Knowledge-Distilled Feature Alignment (KDFA) and Self-Supervised Feature Alignment (SSFA) to guide the output of intermediate feature layer. Without loss of generality, the norm $\|\cdot\|$ is used to measure the distance between two feature vectors, which can be formulated as follows:
	\begin{equation}
		\mathcal{L}_{\text {fea1}} = \|S(x_{adv}) - T(x_{clean})\| _{p},
	\end{equation}
	\begin{equation}
		\mathcal{L}_{\text {fea2}} = \|S(x_{adv}) - S(x_{clean})\| _{p},
	\end{equation}
	where $x_{clean}$, $x_{adv}$, $S(\cdot)$, $T(\cdot)$ are clean images, adversarial images, feature extractor of student and teacher. And $p$ specifies a particular norm. $\mathcal{L}_{\text {fea1}}$ denotes KDFA branch, which guides $S(x_{adv})$ to imitate $T(x_{clean})$. $\mathcal{L}_{\text {fea2}}$ denotes SSFA branch between $S(x_{adv})$ and $S(x_{clean})$, which are generated from student detector. 
	
	$S(x_{clean})$ and $T(x_{clean})$ are served as soft labels to guide the learning of $S(x_{adv})$. Nevertheless, KDFA and SSFA are two independent branches and $S(x_{clean})$ is mutable during the training process. Therefore we can not ensure the credibility of $S(x_{clean})$ in the guiding process. Intuitively, mutable $S(x_{clean})$ will make SSFA task easier to achieve than KDFA. We conjecture the inconsistency of KDFA and SSFA can bring imbalanced competitions during training, which is the main reason why clean precision is lower than standard training in Vanilla-FA. Moreover, inconsistent soft labels might hurt the distillation performance, leading to further potential conflicts and sub-optimal results.
	
	\begin{figure}[t]
		\centering
		
		\includegraphics[width=0.9\linewidth]{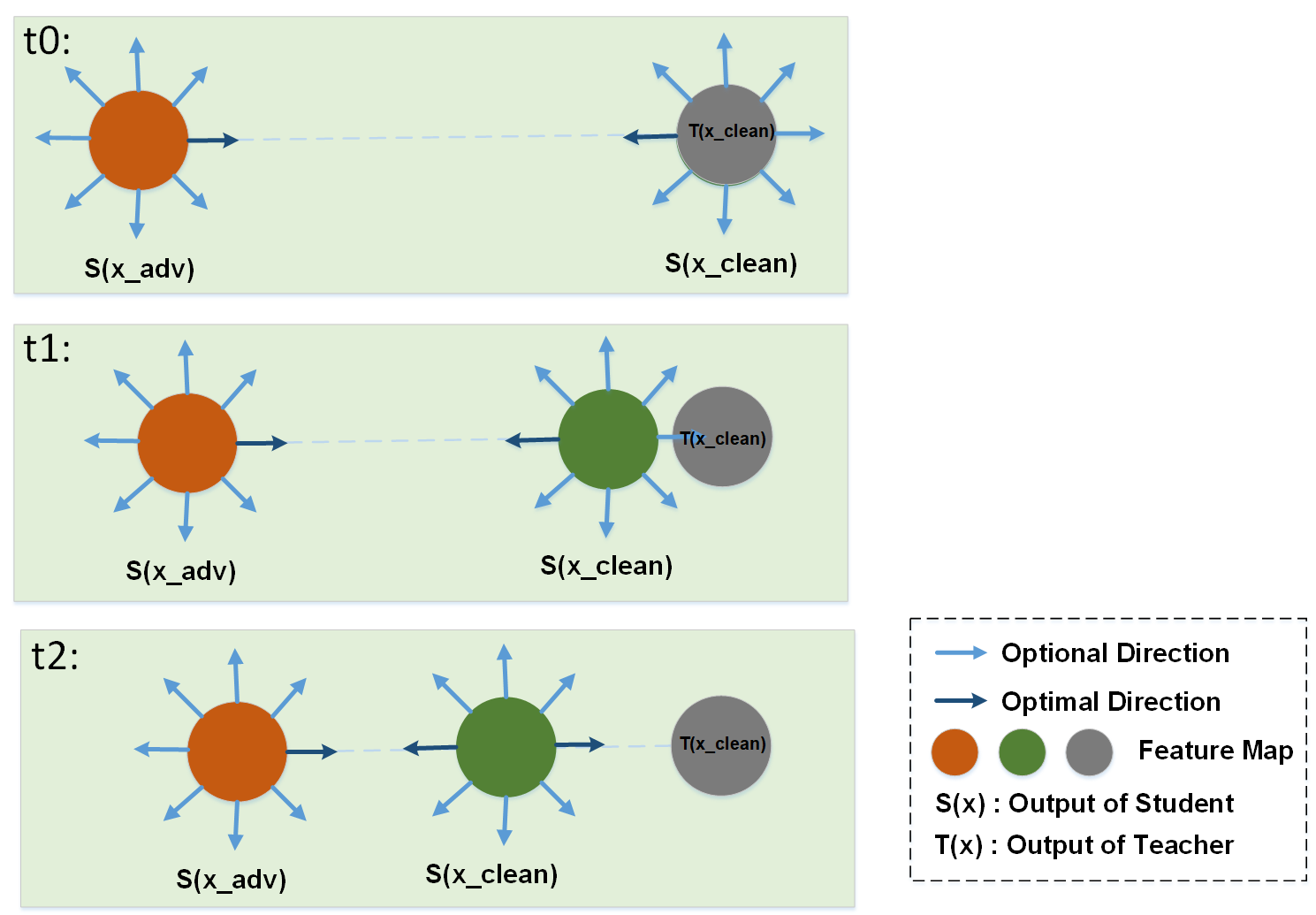}
		
		\caption{The states of feature vectors from different source during the training process.}
		\label{fig:feature_space}
	\end{figure}
	
	\subsection{Decoupled Knowledge-Distilled Feature Alignment Branches}
	\label{sec:theory}
	
	We propose decoupled knowledge-distilled feature alignment branches to avoid potential conflict between vanilla KDFA and SSFA branch. Following the triangle inequality, we can disentangle vanilla KDFA branch into two sub-branches, which can be expressed as follows:
	\begin{equation}
		\mathcal{L}_{\text {fea3}} = \|S(x_{clean}) - T(x_{clean})\| _{p},
	\end{equation}
	\begin{equation}
		\begin{split}
			\mathcal{L}_{\text {fea1}}
			\leq \mathcal{L}_{\text {fea2}} + \mathcal{L}_{\text {fea3}}
		\end{split},
		\label{equ1}
	\end{equation}
	where $\mathcal{L}_{\text {fea3}}$ denotes a new KDFA branch between $S(x_{clean})$ and $T(x_{clean})$. In feature alignment framework, initial state can be expressed as subfigure t0 in Figure \ref{fig:feature_space} when we use the teacher detector to initialize student detector. In that case $S(x_{clean})$ and $T(x_{clean})$ are completely coincident in feature space. When we minimize $\mathcal{L}_{\text {fea2}}+ \mathcal{L}_{\text {fea3}}$, the optimal directions of $S(x_{clean})$ and $S(x_{adv})$ can be determined, as shown in subfigure t0 and subfigure t1 of Figure \ref{fig:feature_space}. Finally, $\mathcal{L}_{\text {fea2}}+ \mathcal{L}_{\text {fea3}}$ will converge in a balanced state in total loss function, as shown in subfigure t2 of Figure \ref{fig:feature_space}. During the training process, $S(x_{adv}) - S(x_{clean})$ and  $S(x_{clean}) - T(x_{clean})$ are always the same direction in feature space. Specifically, $\mathcal{L}_{\text {fea1}} $ and $\mathcal{L}_{\text {fea2}} + \mathcal{L}_{\text {fea3}}$ are exactly equivalent when $p=1$.
	
	Theoretically, minimizing $\mathcal{L}_{\text {fea1}}$ and minimizing $\mathcal{L}_{\text {fea2}}+ \mathcal{L}_{\text {fea3}}$ have the same optimal solutions, namely $S(x_{adv})$, $S(x_{clean})$ and $T(x_{clean})$ coinciding in feature space. However, the parameters of student model are different from teacher during the training process. Therefore the optimal solutions are almost impossible to obtain under total loss function in detection task. However, when $\mathcal{L}_{\text {fea2}}+ \mathcal{L}_{\text {fea3}}$ or $\mathcal{L}_{\text {fea1}}$ converges in a balanced state, $S(x_{clean})$ and $S(x_{adv})$ will be constrained to the manifold of $T(x_{clean})$ in feature space. \cite{DBLP:conf/cvpr/Stutz0S19} demonstrates that on-manifold adversarial samples can boost generalization capability in classification. Also we argue that decoupled knowledge-distilled feature alignment branches constrains $S(x_{adv})$, $S(x_{clean})$ and $T(x_{clean})$ in the same manifold, which can boost generalization capability across adversarial samples and clean samples in object detection.
	
	\subsection{UDFA Framework}
	Based on the above analysis, UDFA is designed by adding a new KDFA branch between $S(x_{clean})$ and $T(x_{clean})$ based on Vanilla-FA.  The overall architecture of UDFA is illustrated in Figure \ref{fig:overall_framework}. The teacher detector is trained on clean datasets with standard training setting, and served as pretrained model to initialize student in adversarial training. For fair comparisons, we do not use more powerful teacher in our UDFA framework. Therefore, UDFA is essentially a self-distillation process, which aims at exploring more information from pretrained detector. Unlike traditional knowledge distillation, our framework also involves self-supervised learning branch to improve robustness, and aims at addressing the conflict between original knowledge distillation task and self-supervised learning task. $S(x_{clean})$ in $\mathcal{L}_{\text {fea2}}$ will be conducted stop-grad operation inspired by SimSiam \cite{DBLP:conf/cvpr/ChenH21}.  In order to improve the effectiveness of distillation, we adopt decoupled fore/back-ground features \cite{DBLP:conf/cvpr/Guo00W0X021} instead of global pooling feature used in Vanilla-FA. Compare with global pooling feature, decoupled features contains location information by distinguishing foreground and background feature, more friendly to object detection task. The overall training targets can be summarized as: 
	\begin{equation}
		\label{eq:total}
		\begin{aligned}
			&\mathcal{L}_{total} =  \alpha * (\mathcal{L}_{\text {cls}}(x_{clean}) + \mathcal{L}_{\text {loc}}(x_{clean}))  + (1-\alpha) \\ 
			&* (\mathcal{L}_{\text {cls}}(x_{adv})  + \mathcal{L}_{\text {loc}}(x_{adv})) + \beta * \mathcal{L}_{\text {fea}}(x_{clean}, x_{adv}),
		\end{aligned}
	\end{equation}
	where $\mathcal{L}_{\text {loc}}(\cdot)$ is the location loss of detector (e.g., GIoU loss \cite{DBLP:conf/cvpr/RezatofighiTGS019}) and $\mathcal{L}_{\text {cls}}(\cdot)$ is the classification loss (e.g., focal loss \cite{DBLP:conf/iccv/LinGGHD17}). $\alpha$ is the weight to balance the detection loss of clean images and adversarial images. $\mathcal{L}_{\text {fea}}(\cdot)$ is the feature alignment loss in UDFA. $\beta$ controls the trade-off between detection loss and feature alignment loss. 
	
	\vspace{-10pt}
	\paragraph{Feature Alignment Loss.} Feature alignment loss $\mathcal{L}_{\text {fea}}$ is composed of (1) KDFA loss $\mathcal{L}_{\text {fea1}}$ between $S(x_{adv})$ and $T(x_{clean})$; (2) SSFA loss $\mathcal{L}_{\text {fea2}}$ between $S(x_{adv})$ and $S(x_{clean})$; (3) KDFA loss $\mathcal{L}_{\text {fea3}}$ between $S(x_{clean})$ and $T(x_{clean})$. The complete feature alignment loss can be expressed as follows:
	
	\begin{equation}
		\begin{aligned}
			&\mathcal{L}_{\text {fea}}(x_{clean}, x_{adv}) = \lambda *\mathcal{L}_{\text {fea3}}(S(x_{clean}), T(x_{clean})) + \\
			&\mathcal{L}_{\text {fea1}}(S(x_{adv}), T(x_{clean})) + \mathcal{L}_{\text {fea2}}(S(x_{adv}), S(x_{clean})).
		\end{aligned}
	\end{equation}
	
	Specifically, $\lambda$ is equal to $1$ in UDFA framework, and  $\lambda=0$ is the case of Vanilla-FA.
	
	For feature alignment loss, UDFA uses $\ell_2$ norm to measure the distance of feature maps, namely $p=2$. Unlike Vanilla-FA only using the output of backbone, UDFA uses multi-level features from feature pyramid network to transfer dark knowledge, which can provide more powerful supervised information to guide feature alignment. Motivated by \cite{DBLP:conf/cvpr/Guo00W0X021}, $\mathcal{L}_{feai}$ is defined as follows:
	
	\begin{equation}
		\begin{aligned}
			&\mathcal{L}_{feai}(\mathcal{S}, \mathcal{T})=\frac{\gamma_{b g}}{N_{b g}} \sum_{h=1}^{H} \sum_{w=1}^{W} \left(1-M_{h, w}\right)\left(\mathcal{S}_{h, w}-\mathcal{T}_{h, w}\right)^{2} \\
			&+\frac{\gamma_{f g}}{N_{f g}} \sum_{h=1}^{H} \sum_{w=1}^{W} M_{h, w}\left(\mathcal{S}_{h, w}-\mathcal{T}_{h, w}\right)^{2}, \quad {i=1,2,3}
		\end{aligned}
	\end{equation}
	where $\mathcal{S}$ and $\mathcal{T}$ denote feature map from student or teacher. $N_{f g}$ and $N_{b g}$ are the number of elements in foreground regions and background regions. $\gamma_{f g}$ and $\gamma_{b g}$ are trade-off parameters in fore/back-ground features. $M$ is the binary mask of feature map corresponding to input image. $h$ and $w$ denote the size of feature map.
	
	\vspace{-10pt}
	\paragraph{Adversarial Attack in UDFA.} In this paper, we use typical and powerful PGD to generate adversarial samples.  Generally, the $k$-step PGD (PGD-$k$) attack can be expressed as follows:
	
	\begin{equation}
		\label{eq:gen}
		x^{t+1}=\mathcal{P}(x^{t}+\frac{\varepsilon}{k}  \operatorname{sign}(\nabla_{x} (\mathcal{L}_{\text {cls}}(x^{t}) + \mathcal{L}_{\text {loc}}(x^{t})))),
	\end{equation}
	where $\varepsilon$ is the perturbation budget. $k$ is PGD steps, equivalent to the number of iterations to generate final adversarial samples for a given image. $\mathcal{P}$ denotes projection operation $\left\{{x^t} \mid\|x^t-x_{clean}\|_{\infty} \leq \varepsilon\right\}$. $x^0$ is equivalent to $x_{clean}$. During training, we adopt PGD-1 to generate $x_{adv}$. During test, different PGD steps are utilized to evaluate the adversarial robustness of detectors.

	\begin{algorithm}[!t]
		\caption{UDFA with AdvProp}
		\KwInput{Dataset $\mathcal{D}$}
		\KwOutput{Learned model parameter $\theta$}
		\begin{algorithmic}[1]
			\FOR{each training epoch}
			\STATE Sample a random batch $x_{clean}\sim \mathcal{D}$
			\STATE Generate $x_{adv}$ with auxiliary batchnorm based on Eq \eqref{eq:gen}
			\STATE Generate $S(x_{adv})$ with auxiliary batchnorm
			\STATE Generate $S(x_{clean})$ with main batchnorm
			\STATE Compute $\mathcal{L}_{\text {cls}}(x_{adv})$ and $\mathcal{L}_{\text {loc}}(x_{adv})$ with auxiliary batchnorm
			\STATE Compute $\mathcal{L}_{\text {cls}}(x_{clean})$ and $\mathcal{L}_{\text {loc}}(x_{clean})$ with mian batchnorm
			\STATE Compute total loss $\mathcal{L}_{total}$ based on Eq \eqref{eq:total}
			\STATE Perform a step of gradient descent $\min\ \mathcal{L}_{total}$
			\ENDFOR
		\end{algorithmic}
		\label{alg:sche}
	\end{algorithm}
	
	\vspace{-10pt}
	\paragraph{UDFA with AdvProp.} Similar to AdvProp, we use one auxiliary batchnorm to take account of adversarial samples during training, and use the main batchnorm for inference. More concretely, $x_{adv}$, $\mathcal{L}_{\text {cls}}(x_{adv})$,  $\mathcal{L}_{\text {loc}}(x_{adv})$ and $S(x_{adv})$ are computed with auxiliary batchnorm. $\mathcal{L}_{\text {cls}}(x_{clean})$,  $\mathcal{L}_{\text {loc}}(x_{clean})$ and $S(x_{clean})$ are computed with the main batchnorm. The overall framework of UDFA with AdvProp is described in Algorithm \ref{alg:sche}.

	\begin{table*}[!tbp]
		\centering
		\scalebox{0.75}{
			
			\begin{tabular}{c| c| c| c| c| c| c| c}
				\hline
				\multicolumn{2}{c|} {\multirow{2}{*}  {method}} &  \multicolumn{3}{c|} {clean AP} & \multicolumn{3}{c} {adv AP} \\
				\cline{3-8}
				\multicolumn{2}{c|} {} & AP & AP50 & AP75 & AP & AP50 & AP75\\
				
				\hline
				\multirow{7}{*} { GFLV1-Res18 } 
				& PRE         & 50.6        & \textbf{74.1}& 54.8       & 5.0         & 16.3          & 1.9 \\
				& STD  & 49.4 (-1.2) & 71.6 (-2.5) & 53.7 (-1.1) & 5.9 (+0.9) & 18.9 (+2.6) & 2.5 (+0.6)       \\
				& Vanilla-AT  & 48.8 (-1.8) & 70.5 (-3.6) & 52.3 (-2.5) & 10.3 (+5.3) & 25.9 (+9.6) & 6.4 (+4.5) \\
				& TOD-AT    & 49.0 (-1.6) & 70.2 (-3.9) & 53.2 (-1.6) & 10.5 (+5.5) & 25.7 (+9.4) & \textbf{6.9 (+5.0)}\\
				& Vanilla-FA  & 49.4 (-1.2) & 71.4 (-2.7) & 53.5 (-1.3) & 10.8 (+5.8) & 29.3 (+13.0) & 5.9 (+4.0)\\
				& Det-AdvProp  & 51.0 (+0.4) & 73.7 (-0.4) & 55.5 (+0.7) & 7.5 (+2.5) & 22.1 (+5.8) & 3.7 (+1.8) \\
				& UDFA (ours) & \textbf{51.5 (+0.9)} & \textbf{74.1 (+0.0)} & \textbf{56.0 (+1.2)} & \textbf{11.0 (+6.0)}  & \textbf{29.5 (+13.2)} & 5.9 (+4.0) \\
				
				\hline
				\multirow{7}{*} { GFLV2-Res18 } 
				& PRE        & 51.1        & 74.6        & 55.3        & 5.9         & 17.1        & 3.2 \\
				& STD        & 50.1 (-1.0) & 72.5 (-2.1) & 54.3 (-1.0) & 7.2 (+1.3) & 20.4 (+3.3) & 3.9 (+0.7)\\
				& Vanilla-AT  & 48.7 (-2.4) & 70.4 (-4.2) & 52.3 (-3.0) & 10.4 (+4.5) & 25.9 (+8.8) & 6.7 (+3.5) \\
				& TOD-AT      & 48.7 (-2.4) & 70.3 (-4.3) & 52.4 (-2.9) & 10.7 (+4.8) & 26.6 (+9.5) & 7.2 (+4.0)\\
				& Vanilla-FA  & 49.4 (-1.7) & 71.4 (-3.2) & 53.1 (-2.2) & 11.2 (+5.3) & 28.9 (+11.8) & 6.8 (+3.6)\\
				& Det-AdvProp & 51.4 (+0.3) & 73.8 (-0.8) & 55.5 (+0.2) & 8.1 (+2.2)  & 22.7 (+5.6) & 4.6 (+1.4) \\
				& UDFA (ours) & \textbf{52.2 (+1.1)} & \textbf{74.8 (+0.2)} & \textbf{56.3 (+1.0)} & \textbf{12.4 (+6.5)} & \textbf{31.2 (+14.1)} & \textbf{7.6 (+4.4)} \\
				
				\hline
				\multirow{7}{*} { GFLV2-Res34 } 
				& PRE         & 54.9        & 77.6        & 59.7        & 7.2 & 20.2 & 3.9 \\
				& STD         & 53.9 (-1.0) & 75.7 (-1.9) & 58.5 (-1.2) & 8.5 (+1.3) & 23.9 (+3.7) & 4.7 (+0.8)\\
				& Vanilla-AT  & 53.1 (-1.8) & 74.3 (-3.3) & 58.0 (-1.7) & 13.7 (+6.5) & 32.6 (+12.4) & 9.4 (+5.5) \\
				& TOD-AT      & 53.4 (-1.5) & 74.9 (-2.7) & 57.8 (-1.9) & 14.4 (+7.2) & 33.8 (+13.6) & \textbf{10.0 (+6.1)}\\
				& Vanilla-FA  & 53.6 (-1.3) & 75.3 (-2.3) & 58.5 (-1.2) & 14.1 (+6.9) & 33.9 (+13.7) & 9.3 (+5.4)\\
				& Det-AdvProp & 55.8 (+0.9) & 77.3 (-0.3) & \textbf{61.4 (+1.7)} & 10.3 (+3.1) & 27.5 (+7.3) & 6.2 (+2.3) \\
				& UDFA (ours) & \textbf{56.1 (+1.2)} & \textbf{77.8 (+0.2)} & 61.3 (+1.6) & \textbf{14.8 (+7.6)}  & \textbf{35.6 (+15.4)} & 9.9 (+6.0) \\

				\hline
		\end{tabular}}
		\caption{Evaluation results on PASCAL-VOC. PRE is pretrained detector. STD denotes standard training. Our proposed UDFA consistently outperforms existing methods under $AP (IoU=.50:.05:.95)$ and $AP50$ metric.}
		\label{ta:main_result}
	\end{table*}
	
	\section{Experiment}
	\label{sec:exp}
	
	\subsection{Datasets and Metrics}
	
	\paragraph{PASCAL-VOC.} PASCAL-VOC dataset contains 20 object classes, and we adopt the standard ``07+12" protocol. The training set is the union of VOC 2007 trainval and VOC 2012 trainval with a total of 16,551 images. The test set is VOC 2007 test with a total of 4,952 images. 
	
	\vspace{-10pt}
	\paragraph{MS-COCO.} MS-COCO dataset contains 80 object classes. We use train2017 and minitrain2017 \cite{DBLP:conf/eccv/SametHA20} for training. The train2017 set contains a total of 118,287 images. The minitrain2017 set is used for more ablation studies, containing 25K images. For testing adversarial robustness, we use val2017 with a total of 5K images. For testing clean precision, we use val2017 and test-dev2017.
	
	\vspace{-10pt}
	\paragraph{Natural Corruptions Benchmark.}  \cite{DBLP:journals/corr/abs-1907-07484} presents a benchmark for assessing the impact of image quality degradation on the performance of detection models. The benchmark includes 15 types of algorithmically generated corruptions from noise, blur, weather and digital categories. Each type of corruption has 5 severity levels with a total of 75 distinct corruptions. 
	
	\vspace{-10pt}
	\paragraph{Metrics.} We consider typical Average Precision as evaluation metric, i.e., $AP (IoU=.50:.05:.95)$, $AP50$, $AP75$. For natural corrupted robustness, we use mean performance under corruption (mPC) as evaluation metric, which can be expressed as follows:
	\begin{equation}
		\mathrm{mPC}=\frac{1}{\mathrm{~N}_{c}} \sum_{c=1}^{\mathrm{N}_{c}} \frac{1}{\mathrm{~N}_{s}} \sum_{s=1}^{\mathrm{N}_{s}} \mathrm{P}_{c, s},
	\end{equation}
	where $\mathrm{P}_{c, s}$ is the performance with corruption $c$ under severity level $s$. $\mathrm{N}_{c}$ and $\mathrm{N}_{s}$ denote the number of corruptions and severity levels. 
	
	\subsection{Implementation Details}
	
	We conduct experiments across one-stage detectors (e.g. GFLV1 \cite{DBLP:conf/nips/0041WW00LT020} and GFLV2 \cite{DBLP:conf/cvpr/LiW0LT021}) and two-stage detectors (e.g. Faster-RCNN \cite{DBLP:conf/cvpr/LinDGHHB17}) with ResNet \cite{DBLP:conf/cvpr/HeZRS16} backbone of different scales. We use MMDetection \cite{DBLP:journals/corr/abs-1906-07155} as the codebase and provide a modified version for various adversarial training algorithms. All the pre-trained detectors (also served as teacher detectors in our work) fine-tuned from ImageNet \cite{DBLP:conf/cvpr/DengDSLL009} pre-trained classifiers. The experiments of standard training and various adversarial training methods use the same hyper-parameter settings. To be more specific, we use SGD optimizer with initial learning rate, $10^{-2}$, momentum, $0.9$, weight decay $0.0001$. For adversarial training, $\alpha$, $\beta$ are set to $0.5$, $1.0$ respectively. In order to reduce the catastrophic overfitting phenomenon \cite{DBLP:conf/nips/AndriushchenkoF20,DBLP:conf/iclr/WongRK20}, we use a small step size in PGD-1 training following Vanilla-FA \cite{9506689}. The step size is set to $2/255$ for training. In particular, original TOD-AT uses a large step size ($8/255$) and only uses adversarial samples for training. Experiments in \cite{9506689} demonstrate that adding clean samples and using a small step size can largely improve performance over results in the work of Zhang et al. \cite{DBLP:conf/iccv/ZhangW19}. For PASCAL-VOC and minitrain2017 of MS-COCO datasets, we repeat sampling each image three times during each epoch. Unless otherwise stated, we apply standard 1x learning schedule (12 epochs), which decays at $9th$ and $12th$ epochs respectively with decay factor $0.1$.

	\subsection{Comparison with Existing Methods}
	We first verify the effectiveness of our proposed UDFA on typical one-stage detector GFLV1 and GFLV2. Comparison of the results on PASCAL-VOC benchmark are shown in Table \ref{ta:main_result}. For adversarial robustness, we use different PGD steps to generate adversarial samples with a standard perturbation budget setting (8/255) for test, i.e., $\varepsilon = 8$. Without loss of generality, we also report the average adversarial AP under different steps (1,2,4), which is denoted as adv AP. See Appendix \textcolor{red}{A} for the corresponding performance under different PGD steps. We also use standard settings (without adversarial training) to fine-tune pretrained detector for a fair comparison, and denote it as STD.  It is observed that STD obtains lower clean AP and higher adv AP than pretrained detector. According to our previous experience, we will attribute the decline in clean precision to the overfitting phenomenon. However, the improvement on robustness demonstrates that there might be a tradeoff between clean accuracy and robustness even without adversarial training settings. For adversarial training algorithms, similar to the previous observations, Vanilla-AT, TOD-AT and Vanilla-FA will hurt clean average precision than standard training. Det-AdvProp can improve clean average precision and robustness than standard training. We can find that for the $AP$ and $AP50$ metric, our UDFA achieves the best performance across clean dataset and adversarial dataset. For more strict $AP75$ metric, UDFA can achieve better or comparable performance compared with existing methods. In particular, UDFA can improve student by about 1.0 clean AP than teacher detector across GFLV1 and GFLV2. Also UDFA can improve adv AP by at least 6.0 points over teacher. 
	
	\begin{table}[!tbp]
		\centering
		\scalebox{0.7}{
			
			\begin{tabular}{c| c c c| c| c| c| c| c| c}
				\hline
				{\multirow{2}{*}  {model}} & {\multirow{2}{*} {$\lambda$}} & {\multirow{2}{*} {$\gamma_{fg}$}} & {\multirow{2}{*} {$\gamma_{bg}$}} &  \multicolumn{3}{c|} {clean AP} & \multicolumn{3}{c} {adv AP} \\
				\cline{5-10}
				& & & & AP & AP50 & AP75 & AP & AP50 & AP75\\
				
				\hline
				\multirow{7}{*} { GFLV1-R18 } 
				& \multicolumn{3}{c|} {Vanilla-FA} &  49.4 & 71.4 & 53.5 & 10.8 & 29.3 & 5.9 \\
				& 0.0 & 0.0 & 6.0 &  49.8 & 71.6 & 53.8 & 10.6  & 28.7  & 5.8  \\
				& 0.0 & 1.0 & 6.0 &  50.4 & 71.9 & 55.1 & 10.9 & 29.1 & 6.5  \\
				& 0.0 & 1.0 & 8.0 &  50.3 & 72.6 & 54.5 & 10.8 & 29.5 & 6.0  \\
				& 1.0 & 0.0 & 6.0 &  51.1 & 73.3 & 55.9 & 11.1  & 30.3 & 6.0  \\
				& 1.0 & 1.0 & 6.0 &  51.5 & 74.1  & 56.0 & 11.0  & 29.5  & 5.9  \\
				& 1.0 & 1.0 & 8.0 &  51.5 & 74.0 & 55.9 & 11.0  & 30.7 & 6.2 \\
				
				\hline
				\multirow{7}{*} { GFLV2-R18 } 
				& \multicolumn{3}{c|} {Vanilla-FA} &  49.4 & 71.4 & 53.1 & 11.2 & 28.9 & 6.8 \\
				& 0.0 & 0.0 & 6.0 &  50.3 & 72.3 & 54.7 & 11.7 & 29.3  & 7.6  \\
				& 0.0 & 1.0 & 6.0 &  50.8  & 73.0 & 54.7  & 11.8   & 29.9  & 7.5 \\
				& 0.0 & 1.0 & 8.0 &  51.0 & 73.7  & 55.5 & 11.3  & 29.6 & 6.7  \\
				& 1.0 & 0.0 & 6.0 &  51.9 & 74.3 & 56.4 & 12.8  & 31.6  & 8.1  \\
				& 1.0 & 1.0 & 6.0 &  52.2  & 74.8  & 56.3 & 12.4  & 31.2  & 7.6  \\
				& 1.0 & 1.0 & 8.0 &  52.2  & 74.8  & 56.8  & 12.2 & 31.1 & 7.6  \\
				
				\hline
				\multirow{7}{*} { GFLV2-R34 } 
				& \multicolumn{3}{c|} {Vanilla-FA} &  53.6 & 75.3 & 58.5 & 14.1 & 33.9 & 9.3 \\
				& 0.0 & 0.0 & 6.0 &  54.8 & 76.8 & 59.7 & 15.5 & 36.4  & 11.3  \\
				& 0.0 & 1.0 & 6.0 &  54.9  & 76.5  & 59.8  & 15.0   & 35.4  & 10.8 \\
				& 0.0 & 1.0 & 8.0 &  55.1  & 76.6  & 60.0  & 15.5   & 36.2  & 11.2  \\
				& 1.0 & 0.0 & 6.0 &  55.8 & 77.4  & 60.7  & 14.0   & 34.2  & 9.0\\
				& 1.0 & 1.0 & 6.0 &  56.1 & 77.8  & 61.3 & 14.8  & 35.6  & 9.9  \\
				& 1.0 & 1.0 & 8.0 &  56.0  & 77.8 & 60.9  & 14.5   & 34.7  & 10.0  \\

				\hline
		\end{tabular}}
		\caption{Different combinations of hyper-parameters for ablation study. Compared with Vanilla-FA, the decoupled features and decoupled feature alignment branches can obtain improvements under different settings.}
		\label{ta:ablation_result}
	\end{table}

	\subsection{Influence of Hyper-parameters}
	
	We further examine the impacts of different hyper-parameters on clean precision and robustness. There are three main hyper-parameters in the UDFA, $\lambda$, $\gamma_{f g}$ and $\gamma_{b g}$. In order to verify the effectiveness of decoupled features and decoupled branches, we explore different combinations of hyper-parameters for ablation study. Considering that the magnitude of gradients from foreground regions are larger than that from background regions, $\gamma_{b g}$ is larger than $\gamma_{f g}$ to balance gradients in our experiments. 
	
	As shown in Table \ref{ta:ablation_result}, we can find that decoupled fore/back-ground features ($\gamma_{f g}>0$ or $\gamma_{b g}>0$) and the decoupled knowledge-distilled feature alignment branches ($\lambda=1$) can both improve the clean precision than Vanilla-FA. More concretely, when we use default setting $\gamma_{f g}=1$ and $\gamma_{b g}=6$, decoupled features  boosts GFLV1 and GFLV2 by 1.0 to 1.4 clean AP, and decoupled branches can further improve 2.1 to 2.8 clean AP than Vanilla-FA. Specifically, if we only use background feature to distill knowledge, there are also an improvement in clean AP than Vanilla-FA. But combining background feature with foreground feature can get better results. We can also see that the clean AP is not sensitive to different $(\gamma_{f g}, \gamma_{b g})$ pairs in a certain range. 
	
	For robustness, the result of adv AP is shown is Table \ref{ta:ablation_result}, and performance under different PGD steps is shown in Appendix \textcolor{red}{A}. On GFLV1-Res18, decoupled features  can obtain comparable robustness than Vanilla-FA, and overall UDFA can improve robustness by 0.2 adv AP. On GFLV2-Res18, decoupled features  and decoupled branches can consistently improve the robustness than Vanilla-FA. On GFLV2-Res34, although adding decoupled branches lowers accuracy by 0.2 adv AP than only using decoupled features in the default setting $\gamma_{f g}=1$ and $\gamma_{b g}=6$, overall UDFA still improve 0.7 adv AP than Vanilla-FA baseline.

	\subsection{Combine UDFA with AdvProp}
	
	\begin{table}[!tbp]
		\scalebox{0.7}{
			\begin{tabular}{lcp{0.05\textwidth}p{0.05\textwidth}p{0.05\textwidth}p{0.05\textwidth}cc}
				\toprule
				\multirow{2}{*}{method} & \multirow{2}{*}{backbone}  & \multicolumn{3}{c}{clean AP}  & \multicolumn{3}{c}{adv AP} \\
				\cmidrule(lr){3-5}\cmidrule(lr){6-8}
				&&AP & AP50 & AP75 & AP & AP50 & AP75\\
				\midrule
				PRE & Res-18 & 51.1 & 74.6 & 55.3 & 5.9 & 17.1 & 3.2\\
				STD & Res-18 & 50.1 & 72.5 & 54.3 & 7.2 & 20.4 & 3.9\\
				Det-AdvProp & Res-18 & 51.4 & 73.8 & 55.5 & 8.1 & 22.7 & 4.6\\
				UDFA (ours) & Res-18 & \textbf{52.9} & \textbf{75.9} & \textbf{57.6} & \textbf{9.1} & \textbf{24.6} & \textbf{5.1}\\
				\midrule
				PRE & Res-34 & 54.9 & 77.6 & 59.7 & 7.2 & 20.2 & 3.9\\
				STD & Res-34 & 53.9 & 75.7 & 58.5 & 8.5 & 23.9 & 4.7\\
				Det-AdvProp & Res-34 & 55.8 & 77.3 & 61.4 & 10.3 & 27.5 & 6.2\\
				UDFA (ours) & Res-34 & \textbf{57.3} & \textbf{79.0} & \textbf{62.6} & \textbf{10.6} & \textbf{28.4} & \textbf{6.2}\\
				\midrule
				PRE & Res-50 & 53.6 & 76.4 & 58.6 & 7.5 & 22.4 & 3.6\\
				STD & Res-50 & 51.7 & 73.2 & 56.5 & 8.5 & 24.1 & 4.4\\
				Det-AdvProp & Res-50 & 54.2 & 75.1 & 59.1 & 10.9 & 29.6 & 6.3\\
				UDFA (ours) & Res-50 & \textbf{55.8} & \textbf{77.2} & \textbf{60.8} & \textbf{11.4} & \textbf{30.9} & \textbf{6.7}\\
				\bottomrule
		\end{tabular}}
		\caption[]{Verify effectiveness of UDFA with AdvProp on one-stage detector (GFLV2).}
		\label{table:gflv2}
	\end{table}
	
	\begin{table}[!tbp]
		\scalebox{0.7}{
			\begin{tabular}{lcp{0.05\textwidth}p{0.05\textwidth}p{0.05\textwidth}p{0.05\textwidth}cc}
				\toprule
				\multirow{2}{*}{method} & \multirow{2}{*}{backbone}  & \multicolumn{3}{c}{clean AP}  & \multicolumn{3}{c}{adv AP} \\
				\cmidrule(lr){3-5}\cmidrule(lr){6-8}
				&&AP & AP50 & AP75 & AP & AP50 & AP75\\
				\midrule
				PRE & Res-18 & 48.5 & 77.0 & 52.2 & 1.8 & 6.0 & 0.9\\
				STD & Res-18 & 47.4 & 74.0 & 51.9 & 2.0 & 6.8 & 0.7\\
				Det-AdvProp & Res-18 & 48.5 & 75.3 & 52.8 & 3.4 & 10.1 & 1.8\\
				UDFA (ours) & Res-18 & \textbf{50.8} & \textbf{77.3} & \textbf{55.7} & \textbf{3.9} & \textbf{11.0} & \textbf{2.2} \\
				\midrule
				PRE & Res-34 & 52.9 & 78.5 & 58.3 & 3.2 & 9.5 & 1.5\\
				STD & Res-34 & 51.8 & 76.8 & 57.4 & 3.4 & 10.0 & 1.3\\
				Det-AdvProp & Res-34 & 53.6 & 77.8 & 59.4 & 5.8 & \textbf{15.0} & 3.8\\
				UDFA (ours) & Res-34 & \textbf{55.7} & \textbf{79.9} & \textbf{62.1} & \textbf{5.9} & \textbf{15.0} & \textbf{4.1}\\
				\midrule
				PRE & Res-50 & 51.4 & 78.3 & 56.2 & 3.0 & 9.4 & 1.4\\
				STD & Res-50 & 49.9 & 75.7 & 54.5 & 2.8 & 9.3 & 1.0\\
				Det-AdvProp & Res-50 & 51.8 & 76.9 & 56.7 & \textbf{5.3} & 14.6 & 3.0\\
				UDFA (ours) & Res-50 & \textbf{54.0} & \textbf{78.7} & \textbf{60.2} & 5.2 & \textbf{14.7} & \textbf{3.1}\\
				
				\bottomrule
		\end{tabular}}
		\caption[]{Verify effectiveness of UDFA with AdvProp on two-stage detector (Faster-RCNN).}
		\label{table:rcnn}
	\end{table}
	
	Based on UDFA framework, we further explore the effectiveness of combining UDFA with AdvProp. To verify this, we conduct more experiments on GFLV2 and Faster-RCNN with various backbones. The results of GFLV2 are presented in Table \ref{table:gflv2}. It is observed that the version of UDFA with AdvProp improves the clean precision by $+1.5\sim+1.6$ AP than SOTA algorithm Det-AdvProp. For robustness, our approach improves by $+0.3\sim+1.0$ AP.
	
	For Faster-RCNN, considering that Det-AdvProp does not conduct experiments on two-stage detectors, we first extend Det-AdvProp to Faster-RCNN. Det-AdvProp disentangles classification and localization loss to generate adversarial samples for training. Intuitively, we set $\mathcal{L}_{\text {loc}}$ consisting of $\mathcal{L}_{\text {roi-loc}}$ and $\mathcal{L}_{\text {rpn-loc}}$, $\mathcal{L}_{\text {cls}}$ consisting of $\mathcal{L}_{\text {roi-cls}}$ and $\mathcal{L}_{\text {rpn-cls}}$. As shown in Table \ref{table:rcnn}, we can find that our approach can improve at least 2.1 clean AP while obtain comparable adversarial robustness.

	\subsection{Results on MS-COCO}
	
	We further conduct experiments on MS-COCO to verify our approach. The pretrained detector is trained with 1x schedule, and other methods are trained with 3x (36 epochs) schedule. The results of GFLV2-Res34 trained on train2017 are summarized in Table \ref{table:train2017}. It is observed that UDFA obtains consistent improvement on val2017 and test-dev2017. For example, UDFA improves clean precision by 1.7 AP over Det-AdvProp on val2017 set, and improves clean precision by 1.9 AP on test-dev2017.  Fore ablation studies, we conduct more experiments on GFLV2 trained with minitrain2017. In Table \ref{table:minitrain2017}, we report the performance on val2017 under different attack strengths. Although there are still large gaps between clean precision and adversarial robustness, our approach achieves better clean precision and robustness than exiting methods.
	
	\begin{table}
		\centering
		\scalebox{0.7}{
			\begin{tabular}{cccccccc}
				\toprule
				method  & dataset &AP & AP50 & AP75 & APl & APm & APs \\
				\midrule
				PRE & val2017 & 39.7 & 56.9 & 43.3 & 52.5 & 43.5 & 22.1\\
				STD & val2017 & 38.6 & 56.2 & 41.6 & 52.0 & 41.1 & 20.6\\
				Det-AdvProp & val2017 & 40.1 & 57.8 & 43.3 & 53.2 & 42.5 & 22.1\\
				UDFA (ours) & val2017 & \textbf{41.8} & \textbf{59.6} & \textbf{45.2} & \textbf{55.7} & \textbf{44.9} & \textbf{23.9}\\
				\midrule
				PRE & test-dev2017 & 40.0 & 57.5 & 43.5 & 50.5 & 42.9 & 21.9\\
				STD & test-dev2017 & 38.9 & 56.5 & 41.9 & 50.2 & 40.7 & 20.7\\
				Det-AdvProp & test-dev2017 & 40.4 & 58.1 & 43.7 & 52.4 & 42.1 & 21.2\\
				UDFA (ours) & test-dev2017 & \textbf{42.3} & \textbf{60.3} & \textbf{46.0} & \textbf{53.9} & \textbf{44.8} & \textbf{23.1}\\
				\bottomrule
		\end{tabular}}
		\caption[]{Clean precision on coco val2017 and test-dev2017.}
		\label{table:train2017}
	\end{table}

	\begin{table}
		\scalebox{0.62}{
			\begin{tabular}{c c c c c c c c c c c}
				\toprule
				method & backbone & 0 & 1 & 2 & 3 & 4 & 5 & 6 & 7 & 8 \\
				\midrule
				STD & Res-18 & 28.3 & 6.6 & 5.8 & 5.4 & 5.2 & 5.2 & 5.2 & 5.2 & 5.2\\
				Det-AdvProp & Res-18 & 28.5 & 6.5 & 5.8 & 5.6 & 5.5 & 5.5 & 5.6 & 5.7 & 5.7\\
				UDFA (ours) & Res-18 & \textbf{31.7} & \textbf{8.8} & \textbf{7.4} & \textbf{7.0} & \textbf{6.8} & \textbf{6.8} & \textbf{6.8} & \textbf{6.9} & \textbf{6.9}\\
				\midrule
				STD & Res-34 & 31.6 & 8.3 & 7.2 & 6.8 & 6.7 & 6.7 & 6.7 & 6.8 & 6.8\\
				Det-AdvProp & Res-34 & 32.2 & 9.2 & 8.2 & 7.9 & 7.9 & 7.9 & 8.0 & 8.1 & 8.2\\
				UDFA (ours) & Res-34 & \textbf{34.6} & \textbf{11.1} & \textbf{9.6} & \textbf{9.1} & \textbf{8.9} & \textbf{8.9} & \textbf{8.9} & \textbf{9.0} & \textbf{9.0}\\
				\midrule
				STD & Res-50 & 31.2 & 8.2 & 7.3 & 7.1 & 7.0 & 7.1 & 7.1 & 7.2 & 7.2\\
				Det-AdvProp & Res-50 & 31.9 & 9.6 & 8.8 & 8.6 & 8.6 & 8.6 & 8.7 & 8.9 & 8.9 \\
				UDFA (ours) & Res-50 & \textbf{34.4} & \textbf{11.0} & \textbf{9.9} & \textbf{9.6} & \textbf{9.5} & \textbf{9.6} & \textbf{9.7} & \textbf{9.8} & \textbf{10.0}\\
				\midrule
				STD & Res-101 & 33.2 & 10.2 & 8.9 & 8.5 & 8.4 & 8.4 & 8.5 & 8.5 & 8.5\\
				Det-AdvProp & Res-101 & 34.0 & 11.9 & 10.7 & 10.4 & 10.3 & 10.4 & 10.4 & 10.5 & 10.6\\
				UDFA (ours) & Res-101 & \textbf{36.9} & \textbf{13.9} & \textbf{12.2} & \textbf{11.7} & \textbf{11.6} & \textbf{11.5} & \textbf{11.5} & \textbf{11.6} & \textbf{11.7}\\
				\bottomrule
		\end{tabular}}
		\caption[]{Evaluation results under different attack strengths.}
		\label{table:minitrain2017}
	\end{table}

	\subsection{Results with Natural Corrupted Augmentation}
	
	\begin{table}
		\scalebox{0.7}{
			\begin{tabular}{lcp{0.05\textwidth}p{0.05\textwidth}p{0.05\textwidth}p{0.05\textwidth}cc}
				\toprule
				\multirow{2}{*}{method} & \multirow{2}{*}{backbone}  & \multicolumn{3}{c}{clean AP}  & \multicolumn{3}{c}{mPC} \\
				\cmidrule(lr){3-5}\cmidrule(lr){6-8}
				&&AP & AP50 & AP75 & AP & AP50 & AP75\\
				\midrule
				PRE & Res-18 & 55.2 & 77.8 & 60.2 & 37.2 & 55.6 & 39.4\\
				STD & Res-18 & 54.6 & 77.0 & 59.5 & 37.1 & 55.1 & 39.2\\
				Det-AdvProp & Res-18 & 55.1 & 77.2 & 60.0 & 39.9 & \textbf{58.9} & 42.4\\
				UDFA (ours) & Res-18 & \textbf{56.8} & \textbf{78.7} & \textbf{62.1} & \textbf{40.0} & 58.6 & \textbf{42.7}\\
				\midrule
				PRE & Res-34 & 58.9 & 80.6 & 64.6 & 40.5 & 59.0 & 43.2\\
				STD & Res-34 & 58.1 & 79.3 & 63.6 & 40.2 & 58.4 & 42.7\\
				Det-AdvProp & Res-34 & 58.2 & 79.4 & 63.6 & 42.9 & 61.6 & 45.9\\
				UDFA (ours) & Res-34 & \textbf{60.3} & \textbf{81.4} & \textbf{66.5} & \textbf{43.2} & \textbf{61.7} & \textbf{46.4}\\
				\midrule
				PRE & Res-50 & 58.7 & 80.5 & 64.4 & 39.7 & 58.2 & 42.3\\
				STD & Res-50 & 57.7 & 79.2 & 63.6 & 39.1 & 57.1 & 41.7\\
				Det-AdvProp & Res-50 & 58.3 & 79.8 & 64.1 & 43.1 & 62.2 & 46.1\\
				UDFA (ours) & Res-50 & \textbf{59.9} & \textbf{81.1} & \textbf{65.7} & \textbf{43.5} & \textbf{62.5} & \textbf{46.6}\\
				\bottomrule
		\end{tabular}}
		\caption[]{Evaluation results on GFLV2 with corrupted augmentation.}
		\label{table:cor}
	\end{table}
	
	\begin{figure}[!htb]
		\centering
		\includegraphics[width=0.95\linewidth]{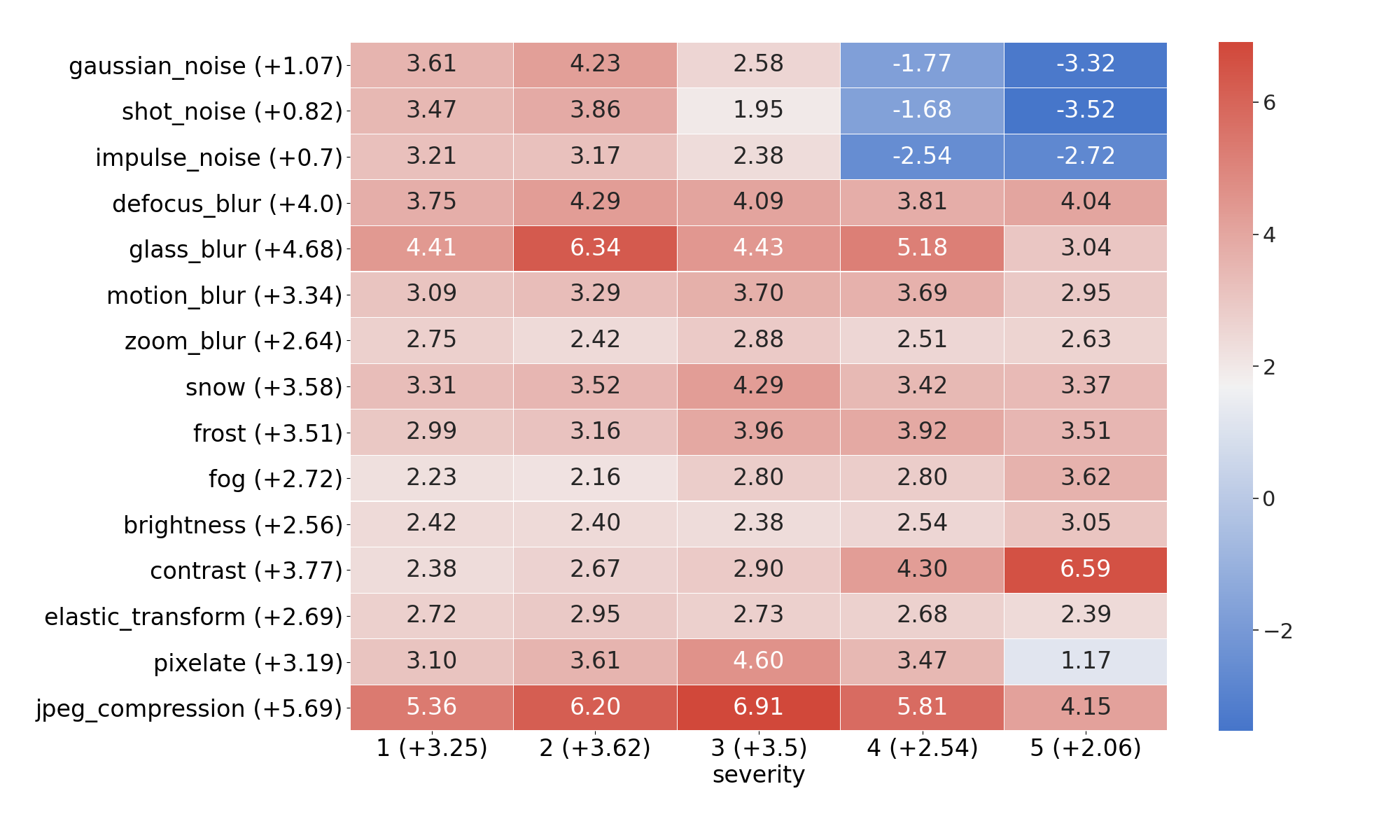}
		\caption{Performance gains of UDFA over standard training on different types of corruptions and severity levels.}
		\label{fig:result_0}
	\end{figure}

	To push the limit of our approach, we further evaluate models on natural corruptions dataset.  We apply the same corrupted augmentation scheme in all methods during training. Specifically, at each iteration, we randomly select one corrupted augmentation from 15 types of corruptions. The severity level is set to 1 during training augmentation. Corrupted augmentation can improve clean AP on standard training and adversarial training methods as shown in Table \ref{table:gflv2} and Table \ref{table:cor}. However, evaluation results of clean AP reveal that Det-AdvProp fails to obtain consistent improvements than pretrained detector with corrupted augmentation. Our approach performs generally well across different backbones and metrics than existing methods even with corrupted augmentation, which further proves the generalization ability of our approach. To further break down the improvement into different types of corruptions, severity levels and classes, the performance gain achieved by UDFA is shown in Figure \ref{fig:result_0} and Appendix \textcolor{red}{B}. We can find that our approach outperforms the standard training on all 15 corruptions and 20 classes on PASCAL-VOC benchmark.  We can observe that the largest improvement when the samples are corrupted by jpeg compression (+5.69 AP).
	
	\section{Conclusion}
	
	In this paper, we propose a new adversarial training framework for building more accurate and robust detector. The proposed UDFA is specifically designed for the fine-tuning process, and benefits from pretrained detector by self-knowledge distillation. From the view of knowledge distillation, we decouple knowledge distillation branch into two sub-branches, which unify knowledge distillation task and self-supervised learning task to avoid potential conflict. Additionally, we adopt decoupled fore/back-ground features for effective distillation and feature alignment. With extensive experiments on PASCAL-VOC and MS-COCO benchmarks, the evaluation results demonstrate that our UDFA can obtain higher clean precision and robustness than standard training, and can obtain higher clean precision and comparable or higher robustness than SOTA adversarial-training-based algorithms.

	\section{Acknowledgement}
	Authors would like to appreciate the Student Innovation Center of SJTU for providing GPUS.

	\clearpage
	
	{\small
		\bibliographystyle{ieee_fullname}
		\bibliography{egbib}
	}
	\clearpage
	
\appendix
\section{Attack under Different Number of PGD Steps}
\label{different_steps}

Comparison with existing methods is shown as Table \ref{tab:a1}. Influence of hyper-parameters is shown as Table \ref{tab:a2}.

\begin{table*}[h]
	\centering
	\scalebox{0.85}{
		\begin{tabular}{c| c| c| c| c| c| c| c| c| c| c}
			\hline
			\multicolumn{2}{c|} {\multirow{2}{*}  {Method}} &  \multicolumn{3}{c|} {PGD-1} & \multicolumn{3}{c|} {PGD-2} & \multicolumn{3}{c} {PGD-4}\\
			\cline{3-11}
			\multicolumn{2}{c|} {} & AP & AP50 & AP75 & AP & AP50 & AP75 & AP & AP50 & AP75\\
			
			\hline
			\multirow{7}{*} { GFLV1-Res18 } 
			&PRE         &  10.6 & 29.3 & 5.5  &  3.0 & 13.3 & 0.3 &  1.3 & 6.3  & 0.0\\
			&STD         &  11.9 & 31.6 & 6.3  &  4.2 & 16.4 & 1.1 &  1.7 & 8.8  & 0.1\\
			&Vanilla-AT  &  19.8 & 38.5 & 16.7 &  6.8 & 22.2 & 1.9 &  4.3 & 16.9 & 0.5\\
			&TOD-AT      &  18.9 & 35.6 & 17.5 &  7.3 & 22.3 & 2.7 &  5.2 & 19.3 & 0.6\\
			&Vanilla-FA  &  19.3 & 41.2 & 15.2 &  7.6 & 25.5 & 1.8 &  5.6 & 21.3 & 0.6\\
			&Det-AdvProp &  15.1 & 36.0 & 10.3 &  5.4 & 19.8 & 0.8 &  2.0 & 10.4 & 0.1\\
			&UDFA (ours) &  19.4 & 41.1 & 15.2 &  8.2 & 26.6 & 1.9 &  5.5 & 20.8 & 0.6\\
			
			\hline
			\multirow{7}{*} { GFLV2-Res18 } 
			&PRE         &  12.3 & 30.7 & 8.4  &  3.5 & 12.7 & 1.0 &  1.8 & 7.8  & 0.1\\
			&STD         &  13.9 & 33.0 & 9.6  &  5.0 & 17.4 & 1.7 &  2.8 & 10.9  & 0.4\\
			&Vanilla-AT  &  18.4 & 35.9 & 16.2 &  7.2 & 22.2 & 2.5 &  5.6 & 19.7 & 1.5\\
			&TOD-AT      &  19.2 & 38.6 & 16.9 &  7.4 & 22.4 & 3.0 &  5.6 & 18.7 & 1.7\\
			&Vanilla-FA  &  20.0 & 41.8 & 15.9 &  7.7 & 24.2 & 2.7 &  5.8 & 20.8 & 1.8\\
			&Det-AdvProp &  16.1 & 37.1 & 11.5 &  5.6 & 20.1 & 1.7 &  2.6 & 11.0 & 0.5\\
			&UDFA (ours) &  21.0 & 43.1 & 17.3 &  9.3 & 27.4 & 3.9 &  6.8 & 23.1 & 1.5\\
			
			\hline
			\multirow{7}{*} { GFLV2-Res34 } 
			&PRE         &  15.1 & 36.2 & 10.2 &  4.3  & 15.6 & 0.8 &  2.2 & 8.7  & 0.6\\
			&STD         &  16.4 & 37.8 & 11.9 &  5.7  & 20.2 & 1.5 &  3.4 & 13.6  & 0.8\\
			&Vanilla-AT  &  22.7 & 44.1 & 20.4 &  10.9 & 29.6 & 5.6 &  7.4 & 24.1 & 2.1\\
			&TOD-AT      &  23.2 & 44.6 & 20.6 &  11.2 & 30.9 & 5.7 &  8.7 & 25.9 & 3.6\\
			&Vanilla-FA  &  23.3 & 46.2 & 20.0 &  11.3 & 30.5 & 5.4 &  7.7 & 25.0 & 2.4\\
			&Det-AdvProp &  20.0 & 43.4 & 15.8 &  7.6  & 24.7 & 2.0 &  3.3 & 14.5 & 0.7\\
			&UDFA (ours) &  22.6 & 45.3 & 19.2 &  12.4 & 33.2 & 6.9 &  9.4 & 28.2 & 3.7\\
			\hline
	\end{tabular}}
	\caption{Comparison with existing methods.}
	\label{tab:a1}
\end{table*}

\begin{table*}
	\centering
	\scalebox{0.85}{
		
		\begin{tabular}{c| c c c| c| c| c| c| c| c| c| c| c}
			\hline
			{\multirow{2}{*}  {model}} & {\multirow{2}{*} {$\lambda$}} & {\multirow{2}{*} {$\gamma_{fg}$}} & {\multirow{2}{*} {$\gamma_{bg}$}} &  \multicolumn{3}{c|} {PGD-1} & \multicolumn{3}{c|} {PGD-2} & \multicolumn{3}{c} {PGD-4}\\
			\cline{5-13}
			& & & & AP & AP50 & AP75 & AP & AP50 & AP75 & AP & AP50 & AP75\\
			
			\hline
			\multirow{7}{*} { GFLV1-Res18 } 
			& \multicolumn{3}{c|} {Vanilla-FA} &  19.3 & 41.2 & 15.2 & 7.6 & 25.5 & 1.8 & 5.6 & 21.3 & 0.6 \\
			& 0.0 & 0.0 & 6.0 &  18.7 & 41.0 & 14.2 & 7.7 & 24.8 & 2.3 & 5.4 & 20.3 & 1.0 \\
			& 0.0 & 1.0 & 6.0 &  19.8 & 41.8 & 16.3 & 7.9 & 25.3 & 2.4 & 5.0 & 20.2 & 0.7 \\
			& 0.0 & 1.0 & 8.0 &  19.6 & 42.2 & 15.7 & 7.7 & 25.8 & 1.8 & 5.2 & 20.5 & 0.6 \\
			& 1.0 & 0.0 & 6.0 &  19.2 & 42.3 & 15.0 & 8.6 & 27.8 & 2.3 & 5.6 & 20.9 & 0.8\\
			& 1.0 & 1.0 & 6.0 &  19.4 & 41.1 & 15.2 & 8.2 & 26.6 & 1.9 & 5.5 & 20.8 & 0.6 \\
			& 1.0 & 1.0 & 8.0 &  19.2 & 41.4 & 15.3 & 8.2 & 26.6 & 2.7 & 5.5 & 21.1 & 0.6 \\
			
			\hline
			\multirow{7}{*} { GFLV2-Res18 } 
			& \multicolumn{3}{c|} {Vanilla-FA} &  20.0 & 41.8 & 15.9 & 7.7 & 24.2 & 2.7 & 5.8 & 20.8 & 1.8 \\
			& 0.0 & 0.0 & 6.0 &  20.2 & 41.5 & 17.1 & 9.0 & 26.4 & 4.2 & 5.8 & 19.9 & 1.6\\
			& 0.0 & 1.0 & 6.0 &  20.3 & 42.5 & 16.8 & 8.9 & 26.3 & 3.7 & 6.2 & 21.0 & 2.1\\
			& 0.0 & 1.0 & 8.0 &  20.0 & 42.1 & 15.7 & 8.2 & 25.1 & 2.8 & 5.8 & 21.6 & 1.6\\
			& 1.0 & 0.0 & 6.0 &  21.7 & 43.4 & 18.7 & 9.2 & 26.9 & 3.6 & 7.5 & 24.6 & 2.1\\
			& 1.0 & 1.0 & 6.0 &  21.0 & 43.1 & 17.3 & 9.3 & 27.4 & 3.9 & 6.8 & 23.1 & 1.5\\
			& 1.0 & 1.0 & 8.0 &  20.5 & 42.8 & 17.1 & 9.2 & 27.4 & 3.8 & 6.8 & 23.1 & 1.9\\
			
			\hline
			\multirow{7}{*} { GFLV2-Res34 } 
			& \multicolumn{3}{c|} {Vanilla-FA} &  23.3 & 46.2 & 20.0 & 11.3 & 30.5 & 5.4 & 7.7 & 25.0 & 2.4 \\
			& 0.0 & 0.0 & 6.0 &  24.6 & 48.1 & 22.1 & 12.8 & 33.1 & 7.7 & 9.1 & 27.9 & 4.1\\
			& 0.0 & 1.0 & 6.0 &  24.4 & 47.4 & 21.9 & 12.3 & 32.4 & 7.4 & 8.3 & 26.3 & 3.1\\
			& 0.0 & 1.0 & 8.0 &  25.6 & 48.6 & 23.2 & 12.3 & 32.8 & 7.1 & 8.7 & 27.1 & 3.4\\
			& 1.0 & 0.0 & 6.0 &  21.9 & 44.2 & 18.8 & 11.8 & 32.2 & 5.5 & 8.2 & 26.2 & 2.7\\
			& 1.0 & 1.0 & 6.0 &  22.6 & 45.3 & 19.2 & 12.4 & 33.2 & 6.9 & 9.4 & 28.2 & 3.7\\
			& 1.0 & 1.0 & 8.0 &  22.5 & 45.0 & 19.6 & 12.0 & 32.0 & 6.6 & 9.0 & 27.0 & 3.7\\
			\hline
	\end{tabular}}
	\caption{Influence of hyper-parameters.}
	\label{tab:a2}
\end{table*}



\section{Results with Natural Corrupted Augmentation}
\label{different_classes}

Performance gain of UDFA over standard training on different classes is shown as Figure \ref{tab:a1}.

\begin{figure*}[!htb]
	\centering
	\includegraphics[width=0.95\linewidth]{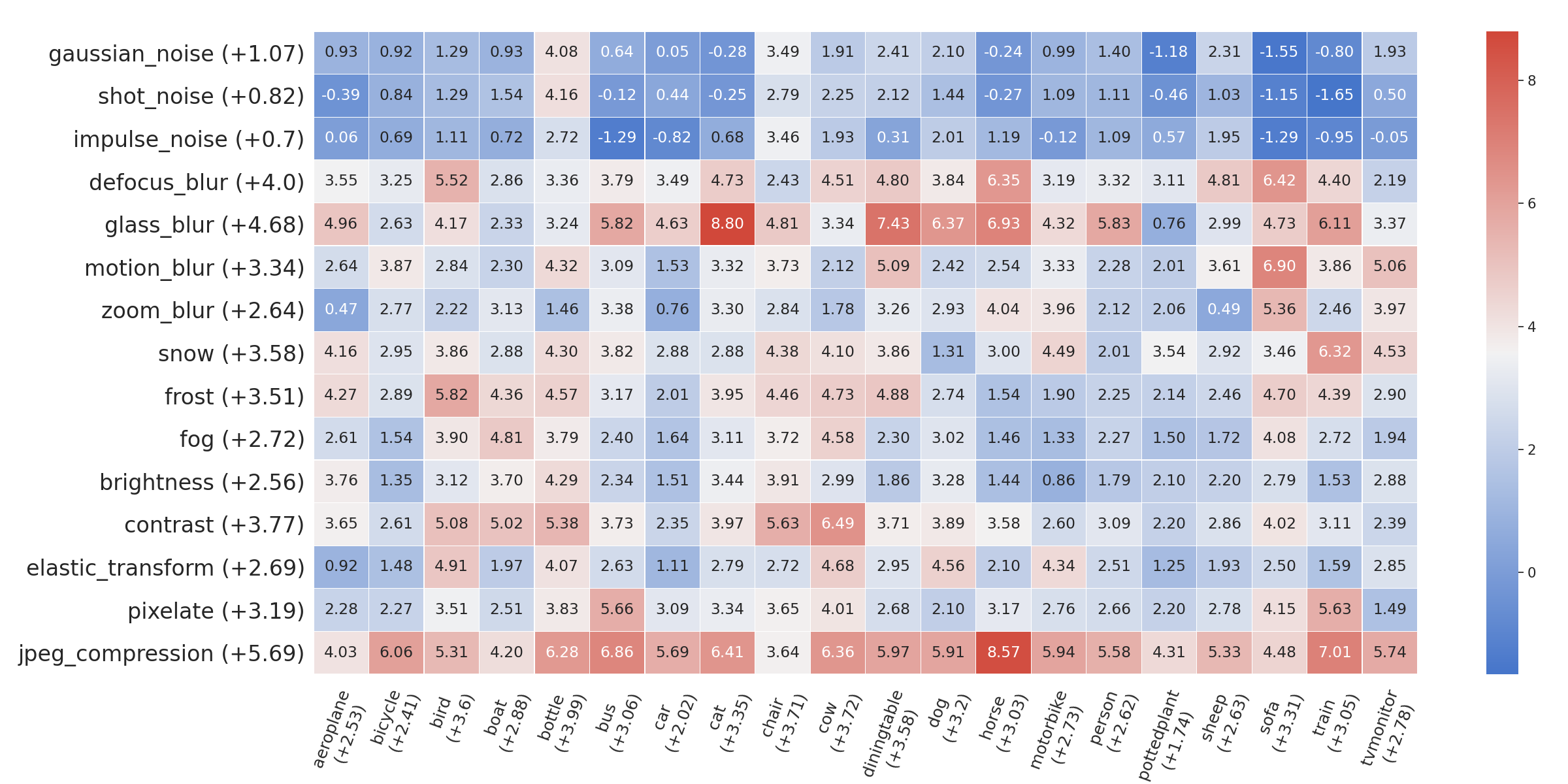}
	\caption{Performance gains of UDFA over standard training on different classes.}
	\label{fig:result_1}
\end{figure*}

\end{document}